\documentclass[10pt,twocolumn,letterpaper]{article}

\PassOptionsToPackage{table}{xcolor}
\usepackage{cvpr}

\usepackage{multirow}

\definecolor{tabfirst}{rgb}{1, 0.7, 0.7}
\definecolor{tabsecond}{rgb}{1, 0.85, 0.7}
\definecolor{tabthird}{rgb}{1, 1, 0.7}

\def\OurMethod{Lift4D\xspace}
\newcommand{\I}{\mathbf{I}}

\newcommand\blfootnote[1]{\begingroup\renewcommand\thefootnote{}\footnote{#1}\addtocounter{footnote}{-1}\endgroup}

\AtEndPreamble{
  \crefname{figure}{Fig.}{Figs.}
  \Crefname{figure}{Figure}{Figures}
  \crefname{equation}{Eq.}{Eqs.}
  \Crefname{equation}{Equation}{Equations}
  \crefname{appendix}{Appx.}{Appxs.}
  \Crefname{appendix}{Appendix}{Appendices}
}

\definecolor{cvprblue}{rgb}{0.21,0.49,0.74}
\usepackage[pagebackref,breaklinks,colorlinks,allcolors=cvprblue]{hyperref}

\def\paperID{}
\def\confName{CVPR}
\def\confYear{2026}

\title{Lift4D: Harmonizing Single-View 3D Estimation \\ for 4D Reconstruction In-the-Wild}

\author{Yehonathan Litman \quad Xiaoxuan Ma \quad Manan Shah \quad Nicol\'as Ugrinovic\\
Kris Kitani$^{*}$ \quad Fernando De la Torre$^{*}$ \quad Shubham Tulsiani$^{*}$\\[3pt]
Carnegie Mellon University\\[3pt]
{\tt\small \url{https://lift4d.github.io}}}

\begin{document}
\twocolumn[{
    \renewcommand\twocolumn[1][]{#1}
    \maketitle
    \begin{center}
    \captionsetup{type=figure}        \vspace*{-0.8cm}
\includegraphics[width=0.88\linewidth,trim=0 1.38cm 0 0,clip]{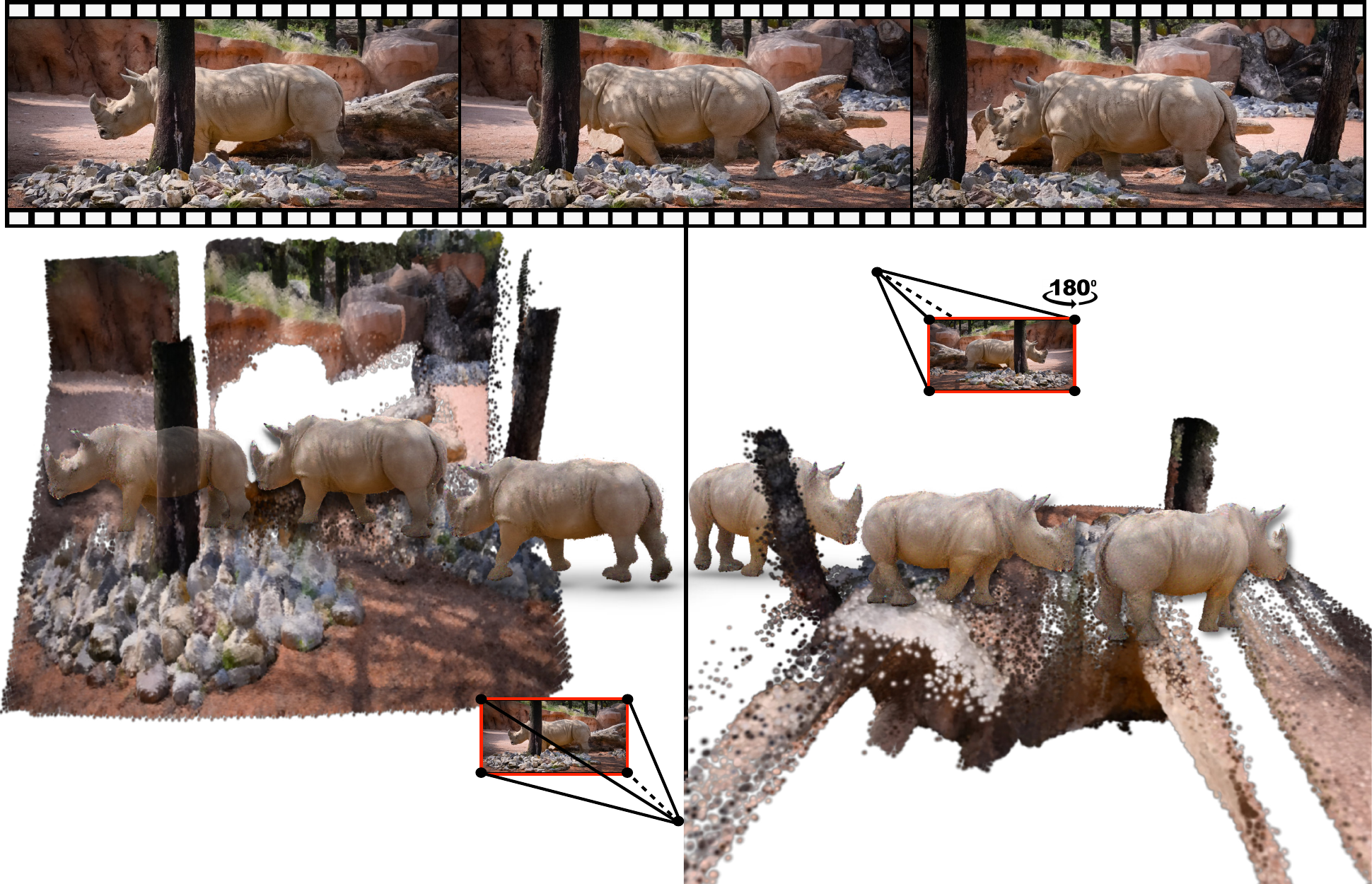}
    \vspace*{-0.3cm}
    \captionof{figure}{
    \textbf{4D Reconstruction from Monocular In-the-Wild Video.} Given a video of a dynamic scene, \OurMethod recovers the full geometry, appearance, and deformation of objects, including regions never observed by the camera, by leveraging a causally conditioned image-to-3D prior and occlusion-aware optimization. The resulting 4D representation handles large deformations and scene occlusions.
    }
    \vspace*{-0.2cm}
    \label{fig:teaser}
    \end{center}
 }]
\maketitle
\blfootnote{$^{*}$Equal co-advising.}

\begin{abstract}
    Reconstructing dynamic non-rigid objects from monocular video requires integrating visual cues from direct observations with data-driven priors over geometry and appearance. Prior approaches either learn to directly predict 4D representations from visual input or initialize a 3D representation that is subsequently deformed and refined based on video evidence. However, the former are constrained by the scarcity of 4D training data, while the latter leverage priors only for the initial reconstruction and rely solely on video supervision thereafter; neither handles complex in-the-wild scenarios with large deformations and occlusions well. We present Lift4D, a test-time optimization framework that addresses both limitations. First, we adapt an existing single-view 3D reconstruction model to yield temporally consistent per-frame predictions via causal latent conditioning, providing a coherent initialization for a deformable 3D Gaussian Splatting representation. We then ``sculpt'' this representation to match the input video through an occlusion-aware optimization that faithfully recovers visible surface details while completing unobserved regions using a view-conditioned diffusion prior. We demonstrate that Lift4D clearly improves over prior 4D reconstruction methods, particularly on challenging in-the-wild sequences with severe occlusions and non-rigid motion.
\end{abstract}

\section{Introduction}
\label{sec:intro}

Consider the video of the rhino in \cref{fig:teaser}.
Despite seeing it from only a handful of viewpoints, we can naturally perceive it as a single, persistent 3D object—mentally completing its unseen surfaces and effortlessly tracking how its shape deforms over time.
This remarkable ability to infer a full, coherent world from partial observations motivates our work.
In this work, we aim to develop a computational method for inferring a complete 4D reconstruction of generic objects from monocular in-the-wild videos: given a single video, we seek to recover both the full 360\textdegree\ geometry and appearance of each dynamic object, along with its deformation across frames.

Inferring 4D representations from monocular input is an open problem that existing approaches address only partially. In-the-wild objects are unconstrained in category, may undergo large deformations, and suffer from occlusions, all compounded by the fundamental ambiguity of a single viewpoint. Addressing these challenges requires leveraging data-driven priors, as purely geometric cues are insufficient for complete 4D reconstruction. Existing approaches face two fundamental limitations. First, methods that directly predict 4D representations~\cite{ren2024l4gm,sabathier2026actionmeshanimated3dmesh,chen2026motion3to4,ShapeGen4D,sun2024eg4d} are bottlenecked by the scarcity of diverse 4D training data: they either depend on category-specific templates~\cite{yang2022banmo,yang2023ppr}, restricting them to narrow object domains, or train on synthetic assets that lack the diversity needed for in-the-wild generalization. Second, optimization-based methods~\cite{jiang2024consistent4d,dreamscene4d,wu2024sc4d} sidestep this by relying on more widely applicable 3D priors, but struggle to bridge the gap between such static priors and dynamic sequences: those leveraging video diffusion priors~\cite{ren2025gen3c,chen2025cognvs,trajectorycrafter} degrade under large viewpoint changes, while those using image-to-3D priors only for initialization~\cite{pad3r,v2m4} suffer from a domain gap between static priors and dynamic sequences, leading to degraded geometry, motion, or appearance under large deformations and occlusions.

Our key insight is that a state-of-the-art single-view 3D reconstruction method (\eg SAM3D~\cite{sam3dteam2025sam3d3dfyimages}) can be adapted to provide strong \emph{4D} priors during optimization. While naively reconstructing each video frame independently yields temporally inconsistent geometry, we introduce a \emph{causal latent conditioning} strategy that makes these per-frame 3D reconstructions temporally coherent by propagating latent information across frames. Nevertheless, such representations remain per-frame and do not form a coherent 3D structure undergoing deformation over time.
To address this, we introduce a \textit{time-varying deformable 3D representation} parameterized by sparse control nodes, which is optimized using the enhanced temporally consistent per-frame reconstructions. To align the deformations with the input video, we also employ rendering-based photometric supervision.
Since in-the-wild videos often contain complex occlusions and unobserved regions, resulting in incomplete supervision, we further propose an \textit{occlusion-aware rendering supervision} scheme. This scheme localizes occluded object regions using depth cues and performs color matching to harmonize the invisible appearance with visible image regions, producing a clean reference image for supervision. Additionally, we utilize generic image diffusion priors~\cite{liu2023zero} to guide the reconstruction of plausible appearances in both occluded and unobserved regions.

Together, these designs enable \OurMethod to reliably reconstruct dynamic 4D representations of generic objects from casual in-the-wild videos, even under rapid motion and large deformations, without relying on multi-view data or category-specific templates.
We evaluate our approach on both synthetic benchmarks and challenging in-the-wild videos featuring large non-rigid deformations and severe occlusions. \OurMethod achieves state-of-the-art 4D reconstruction quality, outperforming existing methods in perceptual quality (LPIPS) and semantic fidelity (CLIP score) on the benchmark \cite{jiang2024consistent4d}, and demonstrating substantially better motion accuracy (EPE) on challenging in-the-wild videos.
The resulting 4D representation naturally yields better dense 4D correspondence tracking as an emergent byproduct.

\section{Related Works}
\label{sec:related_works}

\vspace{0.8em}
\noindent\textbf{Dynamic Reconstruction and Tracking from Videos.}
3D Gaussian Splatting (3DGS)~\cite{kerbl20233gaussian} and dynamic extensions such as 4DGS~\cite{Wu_2024_4DGS}, Dynamic 3DGS~\cite{luiten2023dynamic3DGaussians}, and deformable GS variants~\cite{yang2023deformable3dgs,huang2023sc,duisterhof2024deformgs,zhang2025motionblendergaussiansplatting} augment Gaussians with learned deformation fields or canonical-space representations to capture scene dynamics from video.
Monocular reconstruction methods~\cite{shapeofmotion2024,wang2025gflow,liu2025modgs,stearns2024dgmarbles,lei2024mosca} tackle the harder single-view setting; Shape of Motion~\cite{shapeofmotion2024}, for instance, jointly optimizes a canonical 3DGS and per-frame deformations using long-range 2D track supervision, yielding temporally coherent reconstructions across the observed sequence.
Feedforward approaches~\cite{zhang2024monst3r,xu2025geometrycrafterconsistentgeometryestimation,jiang2025geo4d,jin2025stereo4d,st4rtrack2025,sucar2025vdpm,karhade2025any4d,chen2025easi3r,cong2026flow3r,luo20264rc,zhang2025d4rt,depthanything3,xu20254dgtlearning4dgaussian} among others, predict depth, point maps, scene flow, or Gaussians across time in a single pass. Across all these methods, reconstructions are constrained to the camera's field of view: unobserved object surfaces remain empty or distorted, and these approaches do not complete the full 360\textdegree\ geometry and appearance of the dynamic object. In contrast, our work enables coherent completion of both occluded and fully unobserved regions by anchoring view-conditioned 2D diffusion guidance.

\begin{figure*}[!t]
    \centering
    \includegraphics[width=\linewidth,trim=0.5cm 13cm 0.2cm 0]{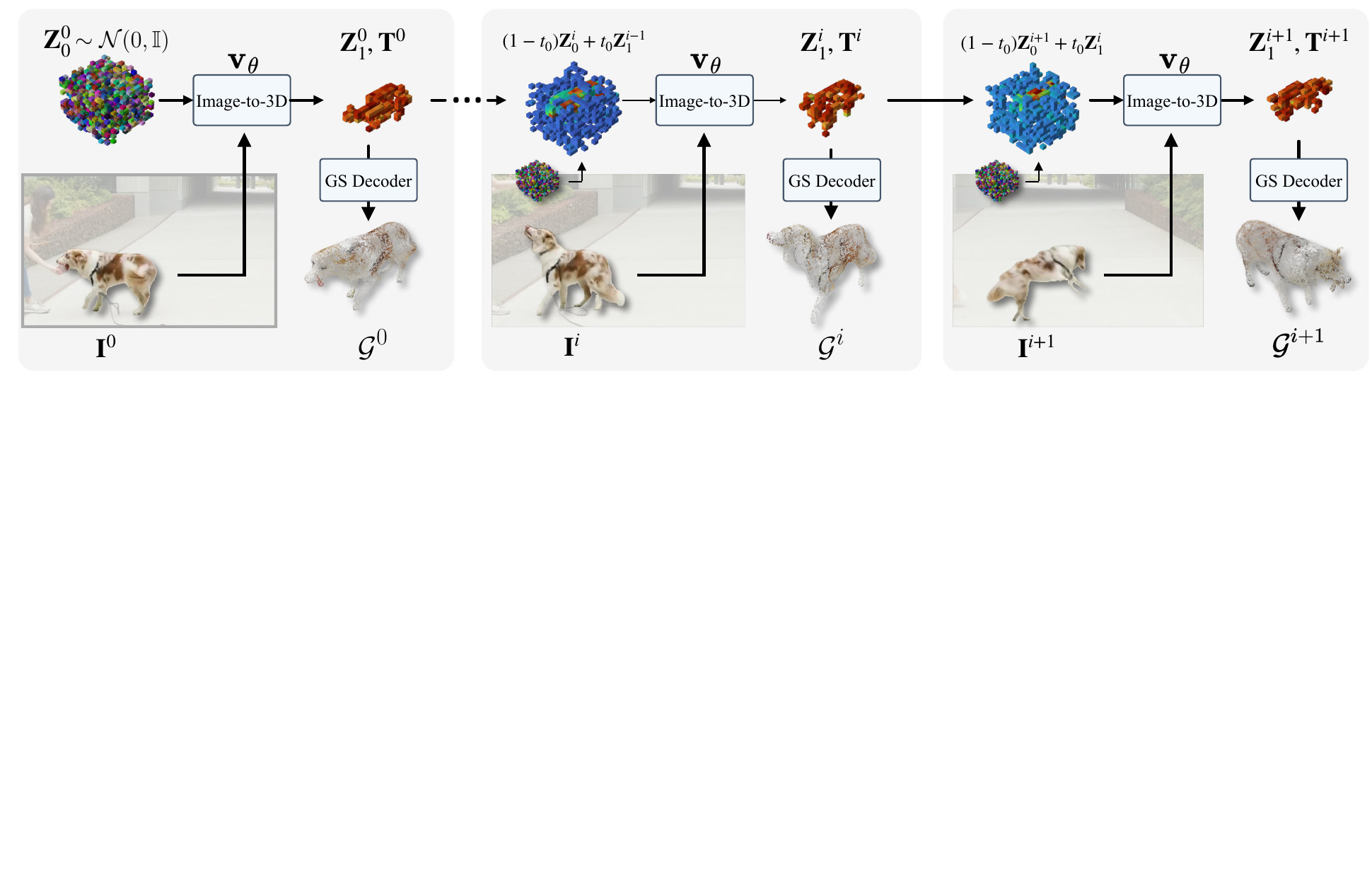}
    \caption{\textbf{Causal Single-view Reconstruction.}
    Given a video input, we obtain per-frame 3D reconstructions $\mathcal{G}^i$ with an image-to-3D model \cite{sam3dteam2025sam3d3dfyimages} using \textit{causal latent conditioning} to enforce the temporal consistency across frames. For a reference frame 0, we first fully denoise a latent $\mathbf{Z}^0_1$ and object-to-camera transform $\mathbf{T}^0$ from a reference canonical frame $\mathbf{I}^0$. The denoised latent is then propagated to the next frame by linearly interpolating it with the next frame's initial noisy latent before beginning the 3D denoising process.
    }
\vspace{-1em}
    \label{fig:approach_sdedit}
\end{figure*}
\noindent\textbf{Generative 4D Novel View Synthesis.} To address viewpoint limitations, a class of methods leverages video diffusion models conditioned on target camera trajectories to hallucinate novel views~\cite{lee2025editbytrack,trajectorycrafter,chen2025cognvs,ren2025gen3c,bai2025recammaster,xie2026lavrscenelatentconditioned}.
These methods ground generation in observed structure via intermediate representations such as depth, point tracks, or geometry latents.
GEN3C~\cite{ren2025gen3c}, for example, projects input frames into an explicit 3D point cloud and uses it to condition a video diffusion model, while CogNVS~\cite{chen2025cognvs} follows a reconstruct, inpaint, then finetune pipeline for dynamic novel-view synthesis from monocular video.
While compelling for moderate viewpoint changes, these methods degrade under extreme extrapolation, where large unseen regions must be hallucinated, owing to the scarcity of diverse multi-view video training data.
Critically, they do not yield an explicit, compositional 4D representation, which limits their utility for downstream applications that require complete and manipulable geometry.

\vspace{0.8em}
\noindent\textbf{Feedforward Generative 4D Reconstruction.}
Rather than generating novel views, another line of work directly predicts complete 4D representations from video in a single forward pass.
L4GM~\cite{ren2024l4gm} trains a large Gaussian reconstruction model on synthetic multi-view video renderings of animated assets, enabling sub-second video-to-4D reconstruction.
ActionMesh~\cite{sabathier2026actionmeshanimated3dmesh} extends 3D latent diffusion with a temporal axis and trains on animated assets~\cite{deitke2023objaverse,deitke2023objaversexl} to produce temporally coherent animated meshes.
Motion 3-to-4~\cite{chen2026motion3to4} decomposes the problem into static shape generation and motion reconstruction, learning compact motion latents over a canonical mesh and predicting per-frame vertex trajectories via a frame-wise transformer.
Further methods pursue related feedforward or diffusion-based backbones~\cite{ShapeGen4D,sun2024eg4d,sabathier2025limlargeinterpolatormodel,zhang2025gaussian}.
Their key limitation is dependence on synthetic or category-specific 4D assets for training, which are expensive to produce and limited in diversity.
Consequently, these models generalize poorly to in-the-wild videos with occlusions, large non-rigid deformations, or novel object categories. Conversely, \OurMethod is not constrained to category-specific templates, addresses occlusions, and can handle large non-rigid deformations.

\vspace{0.8em}
\noindent\textbf{Prior-aided 4D Reconstruction.}
Given the scarcity of 4D training data and multi-view video, a growing body of work builds 4D representations via test-time optimization guided by large-scale 2D or 3D generative priors. One class keeps such a prior continuously in the loop, either as a score-distillation signal over a dynamic Gaussian or NeRF field~\cite{jiang2024consistent4d,dreamscene4d,chu2025generative4dscenegaussiangenmojo,li2024dreammesh4d,zeng2024stag4dspatialtemporalanchoredgenerative,zhang20244diffusion}, or as spatiotemporally consistent multi-view video supervision from a diffusion model~\cite{Wu_2025_cat4d,xie2024sv4d,yao2024sv4d2}. A second class uses a prior only to initialize a canonical geometry from category-specific templates~\cite{yang2022banmo,yang2023ppr} or image-to-3D models~\cite{wu2024sc4d,v2m4}, then refines with video supervision alone; PAD3R~\cite{pad3r}, closely related to our work, initializes a canonical 3D model via an image-to-3D prior, trains a personalized pose estimator on its renderings, and uses the resulting pose initialization to guide deformable Gaussian optimization for category-agnostic reconstruction from casual monocular video. Methods keeping the prior in the loop inherit data scarcity issues or suffer from domain gap due to lacking temporal correspondence, while optimizing from prior-initialized geometry remains ill-posed—many plausible motions and appearances can explain the observed video—often yielding degenerate geometry, motion, or appearance in unobserved regions.
While our work shares the test-time optimization basis of these methods, it addresses their limitations through three components: cross-frame 3D consistency enforcement, explicit modeling of scene-object occlusions, and anchored view-conditioned 2D diffusion guidance.

\section{Methodology}
\label{sec:methodology}

Given a monocular video $\mathcal{I} = \{\mathbf{I}^i\}_{i=1}^N$ with object masks $\mathcal{M} = \{\mathbf{M}^i\}_{i=1}^N$, our goal is to reconstruct a complete 4D representation of individual objects in the scene, factorized into a set of $N_\mathcal{G}$ canonical 3D gaussians and associated deformation parameters.

\begin{figure*}[!t]
    \centering
    \includegraphics[width=\linewidth,trim=0 14.05cm 0cm 0,clip]{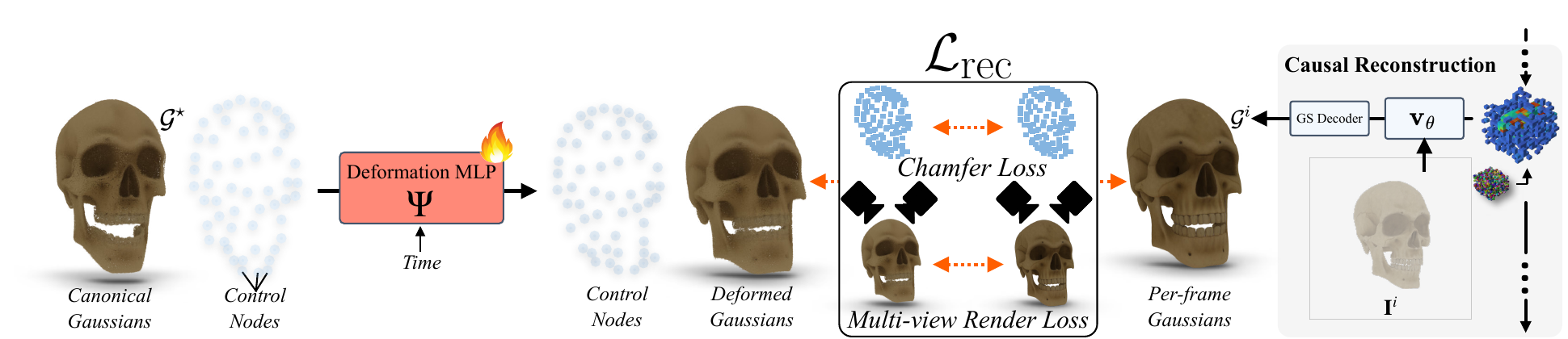}
    \vspace{-2.2em}
    \caption{\textbf{Deformable 3D Optimization.}
    We factorize the 4D representation into canonical 3D gaussians and sparse deformation control nodes and optimize the 4D reconstruction on per-frame reconstructions $\mathcal{G}^i$ via \cref{eq:loss_rec}.}

    \label{fig:approach_deform}
    \vspace{-1.2em}
\end{figure*}

\begin{figure*}[!t]
    \centering
    \includegraphics[width=\linewidth,trim=0cm 14.9cm 5.7cm 0,clip]{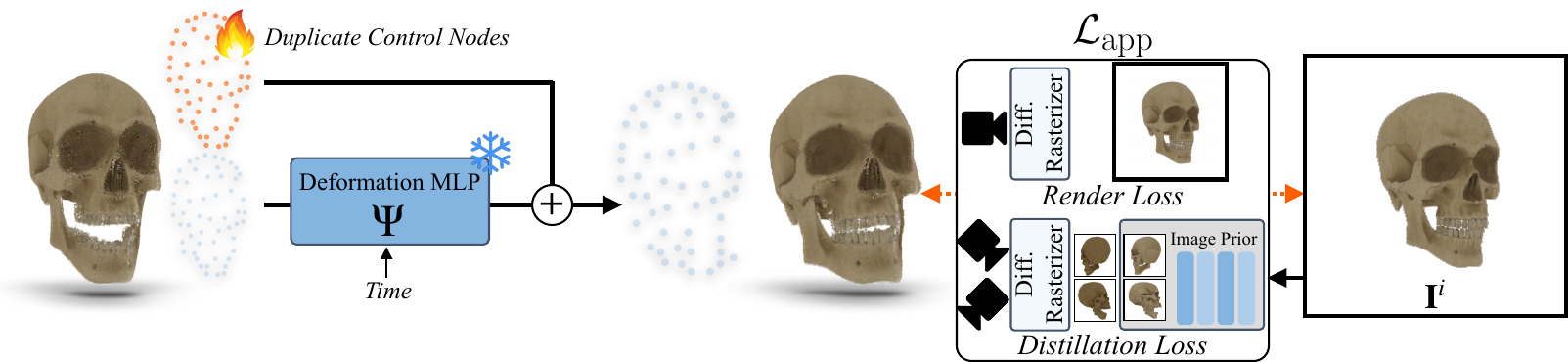}
    \vspace{-2em}
    \caption{\textbf{Appearance Reconstruction.}
    The 3D appearance is deformed with duplicate control nodes and supervised on the reference images $\mathbf{I}^i$ and an image novel view synthesis prior. The reference image supervises observed regions, while the view-conditioned prior supervises unobserved ones via \cref{eq:loss_app}.}
    \vspace{-0.4cm}
    \label{fig:approach_appearance}
\end{figure*}

Monocular in-the-wild videos alone provide far less supervision signal than this representation requires, as most of the object is never fully observed, and the visible portion is often partly occluded. We draw on two large pre-trained 2D and 3D priors and route each to the role where it is reliable for in-the-wild content with a curriculum-based test-time optimization. Off-the-shelf image-to-3D models~\cite{sam3dteam2025sam3d3dfyimages} struggle with appearance fidelity but excel in producing highly detailed geometry; we show they can be adapted to supply a coarse 4D temporally consistent geometric signal (\cref{sec:sdedit}) that is then distilled into a canonical representation (\cref{sec:deform}). Concurrently, view-conditioned image diffusion priors~\cite{liu2023zero} produce inconsistent geometry yet plausible appearance for unobserved views, but by using them only after geometry is fixed, they contribute much higher quality appearance (\cref{sec:occlusion}). By decoupling the priors for geometry and appearance optimization phases accordingly, \OurMethod refines details and infers 4D object reconstructions with consistent geometry and correspondence over time and fine details in visible and occluded regions.

\subsection{Causal Single-view 3D Reconstruction}
\label{sec:sdedit}

We utilize an off-the-shelf flow-matching image-to-3D model ~\cite{liu2022rectifiedflow,sam3dteam2025sam3d3dfyimages} $\mathbf{v}_\theta$ that denoises a structured latent $\mathbf{Z}^i$ encoding geometry and texture in a voxel grid, conditioned on inputs $\mathbf{C}^i$ (image embeddings, metric depth from a monocular depth estimator~\cite{depthanything3}, and the object mask). Applied independently per frame it yields plausible single-view reconstructions but inconsistent geometry across frames. We adapt it into a 4D prior without retraining by coupling adjacent latents at the ODE level, shown in \cref{fig:approach_sdedit}.

\begin{figure*}[!t]
    \centering
    \includegraphics[width=\linewidth,trim=0 6cm 0.4cm 0]{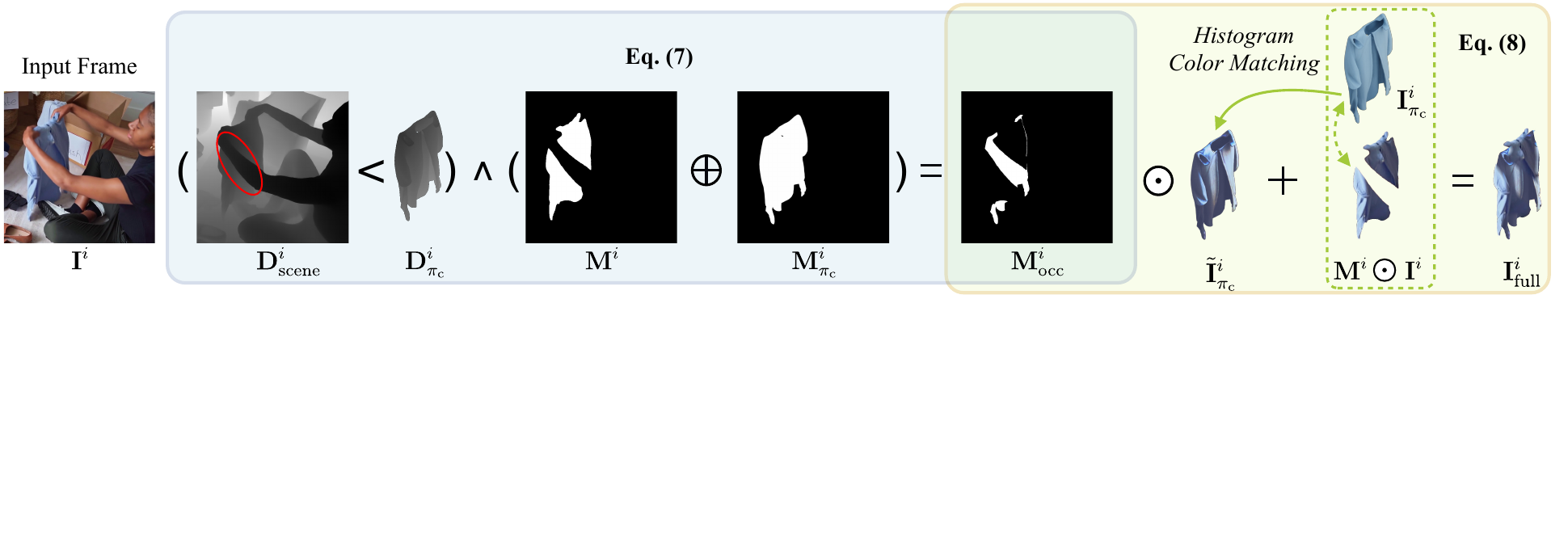}
    \caption{\textbf{Occlusion-aware Rendering Supervision.}
    In cases where the subject is affected by scene occluders, the scene-occlusion mask $\mathbf{M}^i_\text{occ}$ is deduced by comparing the estimated scene depth $\mathbf{D}^i_\text{scene}$ with the rendered object depth $\mathbf{D}^i_{\pi_c}$ (\cref{eq:occ_mask}). The rendered image $\mathbf{I}^i_{\pi_\text{c}}$ is color-matched to the input $\mathbf{I}^i$ over visible regions, producing $\tilde{\mathbf{I}}^i_{\pi_\text{c}}$, which is composited with $\mathbf{I}^i$ into the completed reference $\mathbf{I}^i_\text{full}$ used for supervision (\cref{eq:occ_composite}).}
    \label{fig:occmask}
    \vspace{-1em}
\end{figure*}

\vspace{0.8em}
\noindent\textbf{Causal Latent Propagation.}
We enforce temporal consistency at inference time, without retraining, by reusing the previous frame's denoised latent as a noise prior for the next frame (\cref{fig:approach_sdedit}). We select the first video frame $\mathbf{I}^0$ as the reference frame,
and denoise it from pure Gaussian noise $\mathbf{Z}_0^0 \sim \mathcal{N}(0, \mathbb{I})$ via rectified conditional flow matching:
\begin{equation}
    \mathrm{d}\mathbf{Z}^0 = \mathbf{v}_\theta(\mathbf{Z}^0_t, t, \mathbf{C}^0)\, \mathrm{d}t,
    \label{eq:flow_matching}
\end{equation}
producing a clean structured latent $\mathbf{Z}_1^0$. For each subsequent frame $i$, instead of starting from independent noise, we warm-start the ODE at timestep $t_0 \in (0,1]$ by blending fresh noise with the previous frame's clean latent:
\begin{equation}
\mathbf{Z}^i_{t_0} = (1-t_0)\,\mathbf{Z}^i_0 + t_0\,\mathbf{Z}^{i-1}_1, \qquad \mathrm{d}\mathbf{Z}^i_{t_0} = \mathbf{v}_\theta(\mathbf{Z}^i_{t_0}, t_0, \mathbf{C}^i)\, \mathrm{d}t,
\label{eq:warm_start}
\end{equation}
and integrate from $t_0$ to $1$. The parameter $t_0$ trades temporal consistency against per-frame fidelity, as a larger $t_0$ retains more of the previous frame's structure, while a smaller $t_0$ allows greater per-frame deviation. Propagation runs from the reference frame forward in time. Each denoised latent is decoded by the gaussian splat decoder into per-frame gaussians $\mathcal{G}^i$ with an object-to-camera transform $\mathbf{T}^i \in \mathrm{SE}(3)$ obtained directly from $\mathbf{v}_\theta$.

\subsection{Reconstruction-guided Deformable 3D Optimization}
\label{sec:deform}

The per-frame reconstructions $\{\mathcal{G}^i\}_{i=1}^N$ from \cref{sec:sdedit} are temporally consistent but consist of independent gaussian splat sets without correspondence. We therefore distill them into a deformable canonical representation $\mathcal{G}^\star$ in which the same gaussians explain every frame's 3D reconstruction through a learned deformation (\cref{fig:approach_deform}).

\vspace{0.8em}
\noindent\textbf{Deformable Reconstruction.} We initialize $N_p$ sparse control nodes $\{\mathbf{p}_k\}_{k=1}^{N_p}$~\cite{huang2023sc} on the surface of $\mathcal{G}^\star$, which is initialized from $\mathcal{G}^0$. A deformation MLP $\boldsymbol{\psi}$ predicts each node's time-varying transformation $[\mathbf{R}^i_k|\mathbf{t}^i_k]\in \mathrm{SE}(3)$, to deform every canonical gaussian via linear blend skinning, the details of which we give in the appendix. This sparse parameterization decouples the cost of deformation from the number of gaussians and makes large, non-rigid motions and deformations easy to express when using the causally consistent output. At each iteration we sample a target frame $i$ and minimize a 3D reconstruction loss that aligns the deformed canonical gaussians with the per-frame reconstruction $\mathcal{G}^i$:
\begin{equation}
    \mathcal{L}_\text{rec} = \mathcal{L}_\text{CD} + \mathcal{L}_\text{mv}.
    \label{eq:loss_rec}
\end{equation}

\vspace{0.8em}
\noindent\textbf{Reconstruction Priors.}
The Chamfer term aligns positions while absorbing global per-frame drift via a learnable alignment transform $\mathbf{T}^i_\text{align}$:
\begin{equation}
    \mathcal{L}_\text{CD} = \mathrm{CD}\!\left(\{\boldsymbol{\mu}_m^\star\},\; \{\mathbf{T}^i_\text{align}(\boldsymbol{\mu}_m^{i})\}\right).
    \label{eq:chamfer}
\end{equation}
The multi-view term enforces appearance and depth consistency from a camera $\pi$ randomly sampled on a sphere around the object:
\begin{equation}
    \mathcal{L}_\text{mv} = \mathcal{L}_\text{render}(\hat{\mathbf{I}}^i_\pi, \mathbf{I}^i_\pi)
    \label{eq:random_cam}
\end{equation}
where $\hat{\mathbf{I}}^i_\pi$ and $\mathbf{I}^i_\pi$ are renderings of the deformed gaussians and $\mathcal{G}^i$ respectively, and $\mathcal{L}_\text{render}$ combines $\mathcal{L}_1$ with D-SSIM~\cite{kerbl20233gaussian, huang2023sc}. Together, $\mathcal{L}_\text{CD}$ and $\mathcal{L}_\text{mv}$ tie the deformation to the observed per-frame geometry.

\subsection{Occlusion-aware Appearance Reconstruction}
\label{sec:occlusion}

While the deformable 3D optimization produces a temporally coherent 4D representation, it never directly compares the rendered appearance to the input video. Yet, naively adding a photometric loss against $\mathbf{I}^i$ runs into two distinct problems on in-the-wild sequences. First, when a subject is only partially observed by the input views, pixel supervision is too sparse to constrain geometry and undoes the 3D regularization that $\mathcal{L}_\text{rec}$ already provides. Second, the object may be occluded by surrounding scene content (\eg, the arm covering part of the shirt in \cref{fig:occmask}), so the 4D representation is supervised on incomplete reference pixels even where the image \emph{is} informative. We address these two issues separately. To prevent appearance fitting from corrupting geometry, we freeze the deformation MLP $\boldsymbol{\psi}$ learned in \cref{sec:deform} and add a denser set of control nodes alongside the optimized control nodes, each with its own per-frame $\mathrm{SE}(3)$ deformation. At appearance reconstruction, only the new per-frame transformations and the canonical gaussian attributes are updated, so the coarse motion captured by $\boldsymbol{\psi}$ is preserved while the new nodes absorb the small adjustments needed to fit fine-grained appearance, as shown in \cref{fig:approach_appearance}. To handle occlusion, we combine two complementary supervision signals; a rendering loss that supervises only what is visible, and a diffusion-based image prior that completes non-visible regions with occlusion-completed video images:
\begin{equation}
    \mathcal{L}_\text{app} = \mathcal{L}_\text{render} + \mathcal{L}_\text{SDS}.
    \label{eq:loss_app}
\end{equation}

\noindent\textbf{Occlusion-Aware Rendering.}
We first identify, per frame, which object pixels are occluded by other scene content (\cref{fig:occmask}). For frame $i$, the occlusion mask is
\begin{equation}
    \mathbf{M}^i_\text{occ} = \bigl(\mathbf{D}^i_\text{scene} < \mathbf{D}^i_{\pi_\text{c}}\bigr) \,\wedge\, \bigl(\mathbf{M}^i \oplus \mathbf{M}^i_{\pi_\text{c}}\bigr),
    \label{eq:occ_mask}
\end{equation}
where $\mathbf{D}^i_\text{scene}$ is the monocular scene depth~\cite{depthanything3}, $\mathbf{D}^i_{\pi_\text{c}}$ and $\mathbf{M}^i_{\pi_\text{c}}$ are the depth and alpha mask rendered from $\mathcal{G}^i$ at the input camera $\pi_\text{c}$, $\mathbf{M}^i$ is the SAM3~\cite{carion2025sam} object mask, and $\wedge, \oplus$ denote element-wise AND and XOR. The depth comparison detects pixels where the scene lies in front of the object and the mask XOR restricts attention to foreground object regions. Occluded pixels still need plausible supervision so that the canonical model is not affected by missing data. The most direct source is the per-frame reconstruction $\mathcal{G}^i$, whose rendering is structurally correct but mainly differs in saturation from the input video. We therefore use it only as a color-corrected proxy. Specifically, we compute a per-channel histogram mapping between $\mathbf{I}^i$ and $\mathbf{I}^i_{\pi_\text{c}}$ over visible object pixels (where $\mathbf{M}^i$ is set), apply the mapping to the entire rendered image to obtain $\tilde{\mathbf{I}}^i_{\pi_\text{c}}$, and composite it into the input to form a completed reference view:
\begin{equation}
    \mathbf{I}^i_\text{full} = \mathbf{M}^i_\text{occ} \odot \tilde{\mathbf{I}}^i_{\pi_\text{c}} + \mathbf{M}^i \odot \mathbf{I}^i.
    \label{eq:occ_composite}
\end{equation}
The resulting $\mathbf{I}^i_\text{full}$ uses real video pixels wherever they are trustworthy and falls back to the color-corrected per-frame reconstruction only where there is detected occlusion.

\vspace{0.8em}
\noindent\textbf{Image Priors for Modeling Unobserved Regions.} Even with occlusion handled, $\mathcal{L}_\text{render}$ only supervises visible input view pixels, leaving the rest of the surface unconstrained. To regularize it, we add a score-distillation loss in the spirit of SparseFusion~\cite{zhou2023sparsefusion} using a view-conditioned image diffusion prior~\cite{liu2023zero} conditioned on the occlusion-completed reference. For a randomly sampled novel view $\pi$, we render $\hat{\mathbf{I}}^i_\pi$, encode it via the diffusion encoder to a latent $\mathbf{z}$, sample a timestep $t$ to obtain $\mathbf{z}_t$, denoise it to $\hat{\mathbf{z}}$ with the conditioning $\mathbf{I}^i_\mathrm{full}$, and supervise $\hat{\mathbf{I}}^i_\pi$ against the decoded estimate in pixel space:
\begin{equation}
    \vspace{-0.8em}
    \mathcal{L}_\text{SDS} = \mathbb{E}_{\pi, t} \!\left[ \omega_t \!\left( \|{\hat{\mathbf{I}}}^i_\pi - \mathcal{D}(\hat{\mathbf{z}})\|_2^2 + \mathcal{L}_\text{p}({\hat{\mathbf{I}}}^i_\pi, \mathcal{D}(\hat{\mathbf{z}})) \right) \right],
    \label{eq:sds}
\end{equation}
where $\omega_t$ is a uniform timestep weight and $\mathcal{D}(\cdot)$ is the decoder. Conditioning the prior on $\mathbf{I}^i_\text{full}$ rather than the raw $\mathbf{I}^i$ substantially improves novel-view quality, since the prior is anchored to a clean non-occluded reference. $\mathcal{L}_\text{render}$ and $\mathcal{L}_\text{SDS}$ are therefore complementary, where one supervises the 4D reconstruction on observed pixels while the other hallucinates plausible content in non-visible or occluded regions.

\begin{figure*}[t]
    \centering
    \includegraphics[width=\linewidth,trim=0.1cm 0.8cm 0 0,clip]{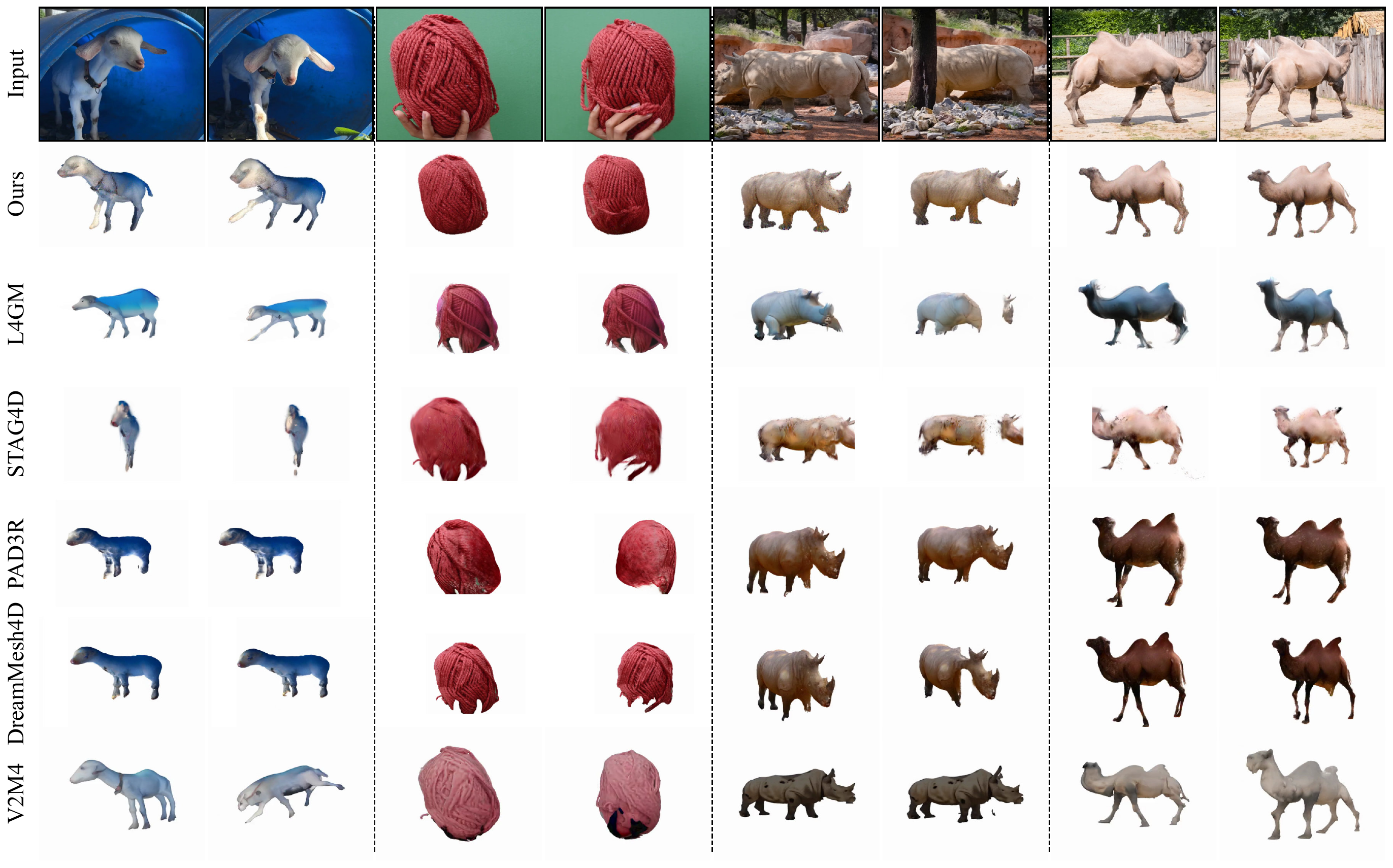}
    \vspace*{-0.6cm}
    \caption{\textbf{Reconstructing 4D Objects from In-the-wild Internet Footage.} 
    Given an input video of an arbitrary object, \OurMethod reconstructs in 4D the complete object geometry and texture. With its usage of a consistent geometry basis and appearance supervision in observed and unobserved regions, our method reconstructs a more topologically accurate geometry and fuses appearance between observed and unobserved regions. On the other hand, the baselines reconstruct badly or show erroneous details in the texture or geometry. Our method works on diverse real-world scenes with large deformations and occlusions.
    }
    \label{fig:inthewild_comparison}
    \vspace*{-0.5cm}
\end{figure*}

\noindent\textbf{Overall Objective.}
The full objective combines the reconstruction, appearance, and structure-prior terms:
\begin{equation}
\mathcal{L} =
\begin{cases}
\mathcal{L}_{\text{rec}} + \mathcal{L}_{\text{reg}}, & k <  N_{\mathrm{rec}} \\
\mathcal{L}_{\text{app}} + \mathcal{L}_{\text{reg}}, & k \geq N_{\mathrm{rec}}
\end{cases}
\end{equation}
where $k$ is the training iteration, $N_{\mathrm{rec}}$ is a predefined number of iterations for 3D optimization, and $\mathcal{L}_\text{reg}$ is a motion regularization term used to regularize the deformation, with the full definition detailed in the appendix.

\section{Experiments}
\label{sec:experiments}

We evaluate \OurMethod's ability to reconstruct complete and temporally consistent 4D representations from monocular video. We compare \OurMethod against other diffusion and feedforward-based 4D reconstruction methods to showcase its effective performance in reconstructing the fidelity, structure, and semantics of the video input when rendered from novel views. This is showcased on synthetic and in-the-wild sequences to demonstrate generalization and real-world applicability. Finally, we ablate core components: the introduced causal temporal conditioning, occlusion-aware video reconstruction, and image prior distillation to demonstrate their necessity for 4D coherence and detail.

\begin{figure*}[!t]
    \centering
    \includegraphics[width=0.999\linewidth,trim=2.5cm 0 0.6cm 0cm,clip]{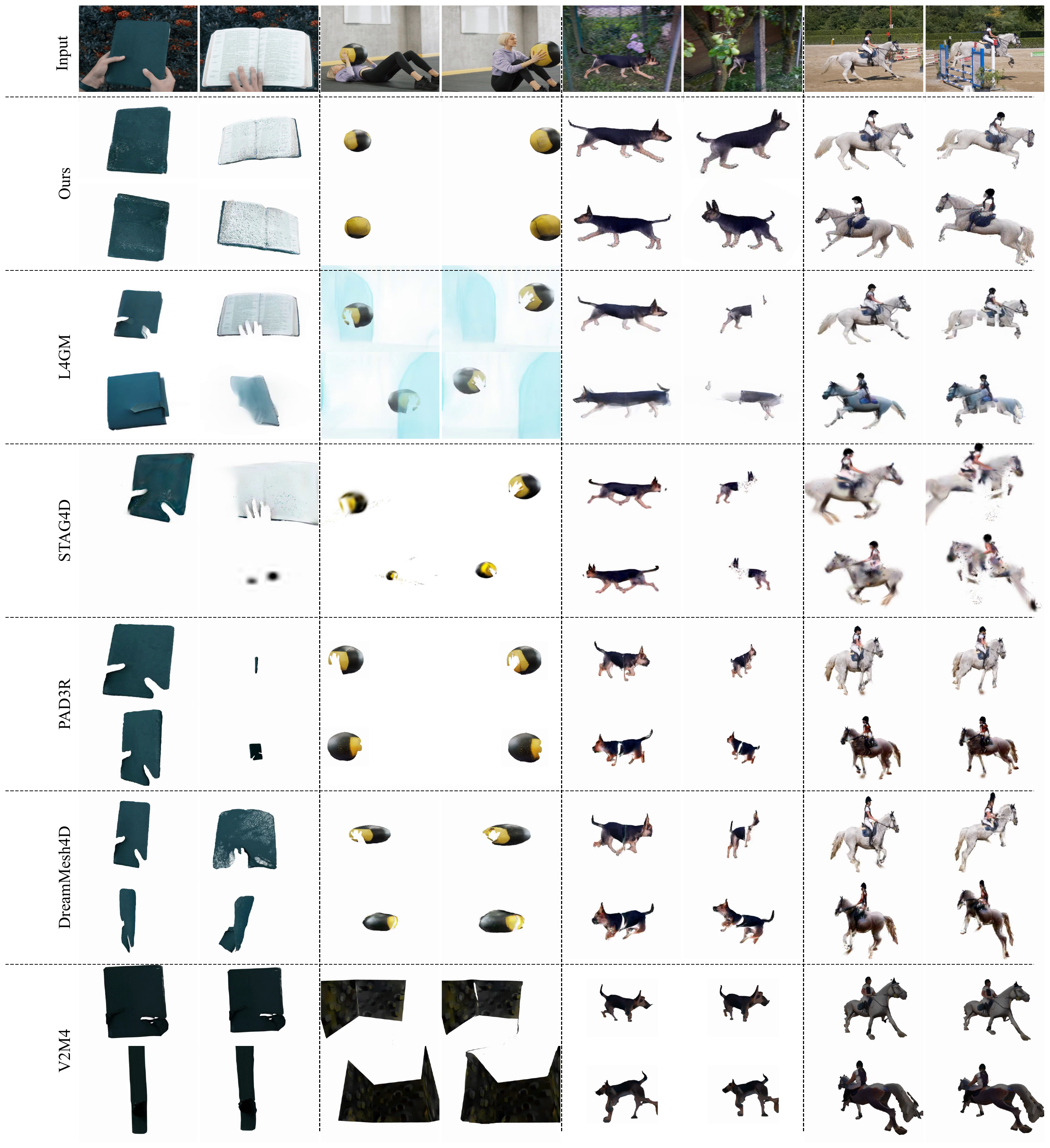}
    \vspace*{-0.5cm}
    \caption{\textbf{Novel Views of 4D In-the-Wild Reconstructions.}     
    We showcase the strong performance of our approach across different novel views on in-the-wild internet stock footage of subjects with occlusions, deformations, and motion. Our method successfully generalizes to multiple types of scenes and objects in-the-wild and their appearance in novel views. The baselines, however, are sensitive to the input and struggle to 4D reconstruct different content consistently across views.
    }
    \label{fig:sup_itw_comparison}
\end{figure*}

\subsection{Experimental Setup}

\vspace{0.8em}
\noindent\textbf{Baselines.}
We compare our approach against other 4D reconstruction baselines \cite{zeng2024stag4dspatialtemporalanchoredgenerative, pad3r, yang2022banmo, ren2024l4gm, li2024dreammesh4d, v2m4} on synthetic and in-the-wild videos that deploy different backbones, \eg diffusion, a feedforward transformer, and test-time optimization using 2D or 3D priors. We first recover the consistent geometry and object-to-camera transforms using SAM 3D \cite{sam3dteam2025sam3d3dfyimages}, and then begin our two-stage test-time optimization, which we run with $N_\mathrm{rec}= 10,000$ for a total of 20,000 iterations with an AdamW optimizer \cite{loshchilov2018adamw}.
Overall, a single object video with 32 frames is reconstructed in {\raise.17ex\hbox{$\scriptstyle\sim$}}30 minutes on one H200 card.

\vspace{0.8em}
\noindent\textbf{Metrics.}
We present qualitative and quantitative comparisons for novel view rendering performance. For synthetic data where we have GT novel view videos, we measure the perceptual similarity with LPIPS, video realism and temporal coherence with Fr\'echet Video Distance (FVD)~\cite{unterthiner2019fvd}, and CLIP score~\cite{hessel2021clipscore} for semantic similarity. For in-the-wild videos where GT novel views are unavailable, we measure the image and text CLIP scores for the predicted novel views and an End-Point Error (EPE) metric~\cite{geng2024motionprompting} to assess the 3D motion accuracy. This is done by measuring the distance between the estimated 3D geometry point tracks projected to the camera view and the GT 2D tracks predicted by CoTracker3~\cite{karaev2024cotracker3}.

\subsection{In-the-Wild 4D Reconstruction}
\label{sec:sup_itw}

\vspace{0.8em}
\noindent\textbf{Dataset.}
For evaluation on in-the-wild videos, we collect a set of 10 publicly available monocular videos from Pexels featuring deformable, rigid, and occluded objects. The videos are characterized by diverse lighting and background conditions and have a subject that can come under scene occlusions at some points. We segment out the subject with SAM 3~\cite{carion2025sam3segmentconcepts} and estimate the scene depth with Depth Anything 3~\cite{depthanything3}. We further include a comparison on 8 real-world videos from DAVIS~\cite{davis_2017}. All videos are between 77 and 100 frames long.

\vspace{0.8em}
\noindent\textbf{Results.}
We show qualitative results on the dataset in Figs.~\ref{fig:inthewild_comparison}-\ref{fig:sup_itw_comparison} along with a comparison to baselines. The results highlight the robustness of our method in accurately recovering 4D reconstructions for diverse scenarios over the baselines, which have difficulty utilizing prior knowledge in real-world scenarios. The rendered novel views show \OurMethod recovers a 4D reconstruction that aligns better with the input view. Furthermore, the quantitative comparisons in Tab.~\ref{tab:sup_itw_full} shows that novel views rendered from our 4D reconstruction are more semantically aligned. The motion of the deformed gaussians reprojected to the camera shows better alignment with CoTracker3 than the baselines, confirming the underlying accuracy of recovered 4D motion.

\subsection{Reconstructing 4D from Synthetic Data}

\vspace{0.8em}
\noindent\textbf{Dataset.}
We first validate \OurMethod on the Consistent4D~\cite{jiang2024consistent4d} 4D synthetic object dataset, which contains 7 input videos of diverse synthetic objects along with 4 novel view videos that are used as GT. All videos consist of 32 frames, and the quantitative comparison is the mean metric across every predicted video for all objects. Scene depth is not computed because the objects are placed by themselves in the center of an empty scene.

\vspace{0.8em}
\noindent\textbf{Results.} As shown in the qualitative comparison in Fig.~\ref{fig:sup_consistent4d_comparison}, \OurMethod excels in recovering asset appearance and geometry in regions unobserved by the camera that are more faithful and topologically accurate compared to the baselines, which show distorted geometry or appearance. This is further reflected by the quantitative comparison in Tab.~\ref{tab:consistent4d_comparison}, showcasing the impact of our design choices on improving novel view semantic and structural quality.

\begin{table}[t]
\centering

\caption{\textbf{Results on In-the-Wild Video Datasets.} Our method greatly outperforms baselines on 4D reconstruction quality and tracking on in-the-wild Pexels data and selected sequences from DAVIS~\cite{davis_2017}.}
\vspace{-0.8em}
\label{tab:sup_itw_full}
\resizebox{.9\linewidth}{!}{
\setlength{\tabcolsep}{2.5mm}
\begin{tabular}{c l c c c}
\toprule
& \textbf{Methods}
& \textbf{CLIP} $\uparrow$
& \textbf{CLIP-T} $\uparrow$
& \textbf{EPE} $\downarrow$ \\
\midrule

\multirow{5}{*}{\rotatebox{90}{\textbf{Pexels}}}

& STAG4D~\cite{zeng2024stag4dspatialtemporalanchoredgenerative}
& \cellcolor{tabsecond} 0.757
& \cellcolor{tabsecond} 0.264
& \cellcolor{tabthird} 0.136 \\

& L4GM~\cite{ren2024l4gm}
& \cellcolor{tabthird} 0.756
& \cellcolor{tabthird} 0.261
& N/A \\

& DM4D~\cite{li2024dreammesh4d}
& 0.705
& 0.256
& 0.140 \\

& PAD3R~\cite{pad3r}
& 0.677
& 0.233
& \cellcolor{tabsecond}0.119 \\

& V2M4~\cite{v2m4}
& 0.745
& 0.256
& 0.211 \\

& \OurMethod
& \cellcolor{tabfirst} 0.780
& \cellcolor{tabfirst} 0.286
& \cellcolor{tabfirst} 0.072 \\

\cmidrule(lr){2-5}

\multirow{5}{*}{\rotatebox{90}{\textbf{DAVIS}}}

& STAG4D~\cite{zeng2024stag4dspatialtemporalanchoredgenerative}
& \cellcolor{tabthird} 0.619
& \cellcolor{tabsecond} 0.270
& \cellcolor{tabsecond} 0.189 \\

& L4GM~\cite{ren2024l4gm}
& 0.600
& 0.235
& N/A \\

& DM4D~\cite{li2024dreammesh4d}
& 0.546
& 0.257
& 0.197 \\

& PAD3R~\cite{pad3r}
& \cellcolor{tabthird} 0.619
& 0.251
& 0.205 \\

& V2M4~\cite{v2m4}
& \cellcolor{tabsecond} 0.637
& \cellcolor{tabthird} 0.263
& \cellcolor{tabthird} 0.195 \\

& \OurMethod
& \cellcolor{tabfirst} 0.715
& \cellcolor{tabfirst} 0.292
& \cellcolor{tabfirst} 0.161 \\
\bottomrule
\end{tabular}}
\end{table}

\begin{table}[t]
\centering
\caption{\textbf{Results on Consistent4D Dataset.} We report the performance of our approach for 4D reconstruction on synthetic object videos against baselines. \OurMethod produces 4D reconstructions that have better structure, semantic quality and coherence compared to the baselines. In each column, the \colorbox{tabfirst}{best}, \colorbox{tabsecond}{second best}, and \colorbox{tabthird}{third best} results are marked.}
\label{tab:consistent4d_comparison}
\vspace{-0.8em}
\resizebox{.9\linewidth}{!}{
\setlength{\tabcolsep}{3.2mm}
\begin{tabular}{l c c c}
\toprule
\textbf{Methods} & \textbf{LPIPS} $\downarrow$ & \textbf{FVD} $\downarrow$ & \textbf{CLIP} $\uparrow$ \\
\midrule
STAG4D~\cite{zeng2024stag4dspatialtemporalanchoredgenerative}
& \cellcolor{tabthird} 0.134
& 1015.57
& 0.917 \\

L4GM~\cite{ren2024l4gm}
& 0.152
& 874.49
& 0.921 \\

DM4D~\cite{dreamscene4d}
& \cellcolor{tabsecond} 0.128
& \cellcolor{tabthird} 688.84
& \cellcolor{tabthird} 0.936 \\

BANMo~\cite{yang2022banmo}
& 0.279
& 1587.10
& 0.808 \\

PAD3R~\cite{pad3r}
& 0.137
& \cellcolor{tabsecond} 645.09
& \cellcolor{tabsecond} 0.942 \\

V2M4~\cite{v2m4}
& 0.192
& 1079.85
& 0.874 \\

\OurMethod
& \cellcolor{tabfirst} 0.116
& \cellcolor{tabfirst} 592.44
& \cellcolor{tabfirst} 0.950 \\
\bottomrule
\end{tabular}}
\end{table}

\subsection{Ablation Studies}

\begin{table}[t]
\centering
\caption{\textbf{Effects of Ablating 3D or Image Priors.} Ablating different 3D heuristic or generative priors in the geometry reconstruction hurts structural quality and coherence on the Consistent4D test set. The image prior is essential for accurately filling in details in unobserved regions.}
\vspace{-0.8em}
\label{tab:ablations}
\resizebox{.9\linewidth}{!}{
\setlength{\tabcolsep}{2.9mm}
\begin{tabular}{l c c c}
\toprule
\textbf{Methods} & \textbf{LPIPS} $\downarrow$ & \textbf{FVD} $\downarrow$ & \textbf{CLIP} $\uparrow$ \\
\midrule
\OurMethod
& \textbf{0.116}
& \textbf{592.44 }
& \textbf{0.950} \\

\OurMethod (Batch-wise $\{\mathcal{G}^i\}_{i=1}^N$)
& 0.120
& 627.90
& 0.945 \\

\OurMethod (No $\mathcal{L}_\text{reg}$)
& 0.122
& 794.82
& 0.943 \\

\OurMethod (No $\mathcal{L}_\text{SDS}$)
& 0.170
& 1242.32
& 0.848 \\

\OurMethod ($\mathcal{L}_\text{rec}$ Only)
& 0.160
& 1079.12
& 0.902 \\

\bottomrule
\end{tabular}}
\end{table}

We run additional optimizations that ablate different physical and image information at optimization such as the initialized geometry, tracking, and velocity motion losses, and distillation from the image prior and evaluate on the Consistent4D dataset. Quantitative comparisons are given in Tab.~\ref{tab:ablations} showing how quality drops when different information and regularization is not provided. Initializing the test-time optimization with $\{\mathcal{G}^i\}_{i=1}^N$ that were inferred batch-wise and thus latent information is independent from one another leads to an overall drop in quality as the deformation quality worsens, causing the geometry to jitter across frames. Excluding $\mathcal{L}_\text{reg}$ also causes the same deformation issues as deformations overfit and jitter across frames. Lastly, not using distillation from the image prior with $\mathcal{L}_\text{SDS}$ leads to a drop in novel view visual quality as optimization relies solely on the initial coarse appearance from $\{\mathcal{G}^i\}_{i=1}^N$, leading to flat and blurry looking appearance in unobserved regions.

\section{Conclusion}
\label{sec:conclusion}

In this paper, we introduced \OurMethod, a test-time optimization framework that successfully recovers complete 4D dynamic objects from monocular video by harmonizing image-to-3D reconstructions as priors for 4D inference. Our approach enables generalizable 4D reconstruction for scenes containing objects with deformations and occlusion interactions by enforcing temporal consistency through causal latent conditioning and utilizing image and 3D priors to refine unobserved regions into a full coherent 4D representation. While \OurMethod significantly improves on state-of-the-art baselines on in-the-wild data, it could be further improved by refining the consistent geometry generation stage in particular. Since it is a cascaded setup and the cascading is controlled via a hyperparameter, performance is inherently tied to the quality of initial SAM3D predictions, and errors can propagate without oversight. Despite these limitations, we believe that improving the underlying architecture's geometry estimation backbone is a promising direction, as further enhancing \OurMethod's generalization to scenes with more complex interaction, such as human grasping, is within possibility.

\paragraph{Acknowledgments.} This work was supported in part by the NSF GFRP (Grant No. DGE2140739) and NSF Award IIS-2345610. This work used Bridges-2 at Pittsburgh Supercomputing Center through allocation CIS240022 from the ACCESS program, which is supported by National Science Foundation grants \#2138259, \#2138286, \#2138307, \#2137603, and \#2138296.

{
    \small
    \bibliographystyle{ieeenat_fullname}
    \bibliography{main}
}

\clearpage
\appendix
\section{Supplementary}
\label{sec:supplementary}

We provide additional quantitative and qualitative results on an expanded dataset of in-the-wild data, as well as comparisons to more baselines. \cref{sec:sup_impl} describes more implementation details of our pipeline.
\cref{sec:sup_metrics} provides definitions of how the CLIP and EPE metrics are utilized for evaluation in the paper and the expanded evaluation. Finally, \cref{sec:sup_limitations} discusses limitations and failure cases.

\subsection{Deformable representation.}
\label{sec:sup_impl}
Given $N_p$ sparse control nodes $\{\mathbf{p}_k\}_{k=1}^{N_p}$~\cite{huang2023sc} on the surface of $\mathcal{G}^\star$. Each node carries a time-varying transformation $[\mathbf{R}^i_k|\mathbf{t}^i_k]\in \mathrm{SE}(3)$, and together they deform every canonical gaussian via linear blend skinning:
\begin{align}
    D^i(\boldsymbol{\mu}_m^\star) &= \sum_{k \in \mathcal{S}} w_{mk}\bigl(\mathbf{R}^i_k(\boldsymbol{\mu}_m^\star - \mathbf{p}_k) + \mathbf{p}_k + \mathbf{t}_k^i\bigr),
    \label{eq:deformation_mu} \\
    D^i(\mathbf{q}_m^\star) &= \Bigl(\sum_{k \in \mathcal{S}} w_{mk}\,\mathbf{q}_k^i\Bigr) \otimes \mathbf{q}_m^\star,
    \label{eq:deformation_q}
\end{align}
where $\mathbf{q}_k^i$ is the quaternion form of $\mathbf{R}_k^i$, $\mathcal{S}$ is the set of $k$ nearest control nodes to $\boldsymbol{\mu}_m^\star$, and the blend weights are
\begin{equation}
    w_{mk} = \frac{\hat{w}_{mk}}{\sum_{k' \in \mathcal{S}} \hat{w}_{mk'}}, \qquad \hat{w}_{mk} = \exp\!\Bigl(\frac{-\|\boldsymbol{\mu}_m^\star - \mathbf{p}_k\|^2}{2 o_k^2}\Bigr),
\end{equation}
with learnable radius $o_k$.

\paragraph{Motion Regularization ($\mathcal{L}_{\text{reg}}$)}

Without regularization, $D^i(\mathcal{G}^\star)$ overfits to per-frame noise in $\mathcal{G}^i$. We therefore add an As-Rigid-As-Possible term~\cite{arap_loss, luiten2023dynamic3DGaussians} $\mathcal{L}_\text{ARAP-GS}$ on the deformed gaussians, which preserves local rigidity, and a control node position smoothness term that penalizes abrupt motion of control nodes:
\begin{equation}
\mathcal{L}_{\text{v-TV}} = \sum_{i=1}^{N-1} \sum_{k=1}^{N_p}
  \|\mathbf{t}_k^{i+1} - \mathbf{t}_k^i\|_2^2.
\label{eq:v_tv_loss}
\end{equation}
where $\mathcal{L}_{\text{reg}} = \mathcal{L}_{\text{v-TV}} + \mathcal{L}_\text{ARAP-GS}$. Together, these priors make the optimization stable under noisy per-frame inputs and enable later stages (\cref{sec:occlusion}) to refine appearance without distorting the geometry.

\subsubsection{Causal Reconstruction.}
For the causal reconstruction described in \cref{sec:sdedit}, we set $t_0 = 0.2$ as the default consistency strength, providing a balance between preserving the previous frame's structure and allowing per-frame deformation. The reference frame $\I^\star$ is set to the first frame. The per-frame object-to-camera transform layout tokens are initialized from $\mathcal{N}(0, \mathbb{I})$.

\subsubsection{4D Optimization.}
We initialize $N_p = 1024$ sparse control nodes on the canonical Gaussian surface using farthest-point sampling and set $k=4$ nearest control nodes per Gaussian for linear blend skinning. We additionally apply Gaussian densification and pruning to adaptively refine the canonical representation. We deploy Stable Zero123~\cite{liu2023zero} as the view-conditioned image diffusion prior for $\mathcal{L}_\text{SDS}$. The diffusion timestep is sampled uniformly from $[0.2, 0.5]$, and the guidance scale is set to $3.0$. All rendering is performed via differentiable 3D Gaussian splatting. For the photometric rendering loss $\mathcal{L}_\text{render}$, we render at the native input resolution. For $\mathcal{L}_\text{mv}$ we render at $512 \times 512$ and for $\mathcal{L}_\text{SDS}$ we crop the input image at $256 \times 256$.

\subsection{Metric Details}
\label{sec:sup_metrics}

\subsubsection{CLIP Score.}
For each method, we render novel orbit views of the reconstructed 4D object at 3 uniformly spaced viewpoints from the input views, i.e., $90^\circ, 180^\circ, 270^\circ$. We compute the cosine similarity between the CLIP embeddings of each rendered view and the corresponding input frame, averaging over all frames and views. This measures how semantically faithful the novel-view reconstructions are to the input video content. For the novel rendered views, we also measure the text alignment score for all views across all sequences. We evaluated on the \texttt{bear, camel, rhino, horsejump-low, horsejump-high, libby, cows, dog} objects in DAVIS \cite{davis_2017}.

\subsubsection{End-Point Error (EPE).}
Since no ground-truth novel views exist for in-the-wild data, we measure motion fidelity in the input camera view. We use CoTracker3 with a grid size of 20 on the input video, producing ${\sim}$400 tracks per video. The CoTracker points at frame 0 are matched to the nearest vertex or gaussian geometry, depending on the method being evaluated, and the predicted trajectory is acquired by tracking the geometry deformation over frames. The two trajectories are compared with EPE.

\subsection{Limitations and Failure Cases}
\label{sec:sup_limitations}

\subsubsection{Dependence on Initial 3D Reconstructions.} Since our pipeline is cascaded, the quality of the final 4D output is dependent on the initial per-frame SAM3D reconstructions and layout predictions. When SAM3D produces poor geometry or layout, these errors propagate to the canonical representation or occlusion-aware frame reconstruction. This is typically the case for videos with high frame-rates and thin objects due to jumps in the per-frame transforms that optimization struggles to account for.

\subsubsection{Balancing for Consistency \& Fidelity.} The conditioning timestep $t_0$ controls the balance between cross-frame consistency and per-frame fidelity. A high $t_0$ can suppress legitimate deformations, while a low $t_0$ may fail to prevent geometric flickering. We use $t_0 = 0.2$ as a default, but note that some sequences may benefit from tuning. Rigid object sequences benefit from using a higher $t_0$.

\newpage

\clearpage
\begin{figure*}[!t]
    \centering
    \includegraphics[width=0.995\linewidth,trim=2.5cm 0 0.6cm 0cm,clip]{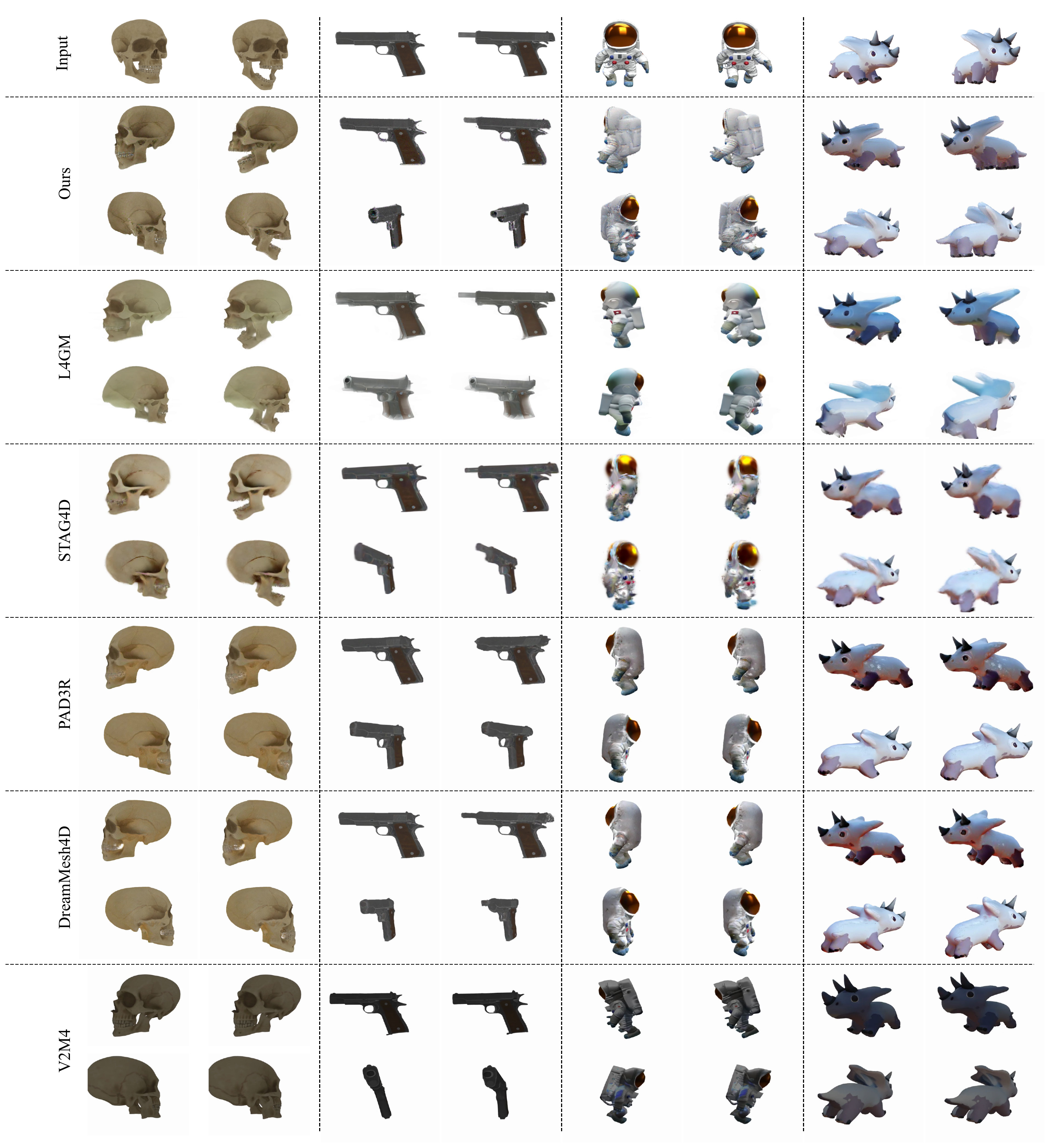}
    \vspace*{-0.2cm}
    \caption{\textbf{Reconstructing 4D Synthetic Objects.} \OurMethod can also robustly reconstruct rich and complete 4D object geometry and texture for simpler synthetic cases, such as in Consistent4D \cite{jiang2024consistent4d}, as opposed to the baselines, which recover simpler geometries and texture or have wrong deformations.}
    \label{fig:sup_consistent4d_comparison}
\end{figure*}
\clearpage

\end{document}